\documentclass[final,3p,times,twocolumn]{elsarticle}

\usepackage{blindtext, graphicx, amsmath, algorithm, algpseudocode, pifont, algcompatible, comment, layout, amsthm, amssymb}
\usepackage{enumitem}
\usepackage{eso-pic}
\usepackage{booktabs}
\usepackage{float}
\usepackage{subfigure}

\usepackage{amsmath,amssymb,amsfonts}
\usepackage{lineno}
\usepackage{graphicx}
\usepackage{textcomp}
\usepackage{algpseudocode}
\usepackage{algorithm}
\usepackage{subfigure}
\usepackage{booktabs}
\usepackage{multirow}
\usepackage{makecell}
\usepackage{epstopdf}
\usepackage{color}
\usepackage{bbding}
\usepackage{multirow}

\usepackage[utf8]{inputenc}
\usepackage[english]{babel}
\usepackage{hyperref}
\hypersetup{ colorlinks=true, linkcolor=black, filecolor=black, urlcolor=cyan, }

\usepackage{caption}
\journal{ICT Express}

\begin{document}

\begin{frontmatter}

\title{Learning-Based Client Selection for Multiple Federated Learning Services with Constrained Monetary Budgets}

\author{Zhipeng Cheng}
\ead{chengzp\_x@163.com}

\author{Xuwei Fan}
\ead{xwfan@stu.xmu.edu.cn}

\author{Ning Chen}
\ead{ningchen@stu.xmu.edu.cn}

\author{Minghui Liwang}
\ead{minghuilw@xmu.edu.cn}

\author{Lianfen Huang\corref{cor1}}
\ead{lfhuang@xmu.edu.cn}

\author{Xianbin Wang}
\ead{xianbin.wang@uwo.ca}

\address{Department of Information and Communication Engineering, Xiamen University, Xiamen, China.\\Department of Electrical and Computer Engineering, Western University, Ontario, Canada.}

\cortext[cor1]{Corresponding author}

\begin{abstract}
We investigate a data quality-aware dynamic client selection problem for multiple federated learning (FL) services in a wireless network, where each client offers dynamic datasets for the simultaneous training of multiple FL services, and each FL service demander has to pay for the clients under constrained monetary budgets. The problem is formalized as a non-cooperative Markov game over the training rounds. A multi-agent hybrid deep reinforcement learning-based algorithm is proposed to optimize the joint client selection and payment actions, while avoiding action conflicts. Simulation results indicate that our proposed algorithm can significantly improve training performance.
\end{abstract}


\end{frontmatter}



\section{Introduction}
Federated learning (FL) represents a promising distributed machine learning paradigm, which enables collaborative model training over a set of client devices, under the coordination of a model aggregator. The implementation of FL-based services in wireless networks has attracted massive attentions \cite{FL-6G}. Different from conventional centralized learning, distributed mobile clients' varying data quality (e.g., data size, data distribution) may impose fundamental impacts on FL performance \cite{DRLFL-CL22}. For example, training data size contributes to the basis of all the learning algorithms \cite{FedIncen-1}, while imbalanced and non-independent and identically distributed (non-IID) data may lead to FL model divergence and degradation \cite{INFOCOM-20,TPDS-22,TMCAuction}. Besides, to fully motivate the clients to participate in FL training, FL service demanders (FLSDs) should pay the clients with monetary rewards as incentives. Therefore, it is critical to achieve efficient trading between the FLSDs and the clients by optimizing the client selection strategy to ensure the benefits of both parties.

Various works have been devoted to the FL client selection problem upon considering different factors. For example, Nishio et al. \cite{FedCS} proposed a client selection protocol to accelerate the FL training process by exploiting the resource conditions of clients. Xu et al. \cite{TWC21} studied the joint client selection and bandwidth allocation problem to ensure long-term FL performance by exploiting clients' current wireless channel information. However, selecting a subset of feasible clients from a large candidate client set poses great challenges due to the high computational complexity, especially when considering the dynamics of clients, e.g., clients with dynamic datasets and resource availabilities. To this end, deep reinforcement learning (DRL)-based methods are widely adopted to solve dynamic client selection problems in wireless networks. Wang et al. \cite{INFOCOM-20} proposed a DRL-based client selection mechanism to increase the learning performance and reduce communication rounds. Jiao et al. \cite{TMCAuction} studied a DRL-based auction framework to incentivize clients to contribute high-quality data to FL services. Deng et al. \cite{TPDS-22} designed a DRL-based quality-aware client selection framework to improve FL efficiency, where the FLSD considers a limited monetary budget for the clients.

Although interesting and meaningful contributions have been made, existing related works only consider single FL service scenarios, while the coexistence of multiple FL services is generally ignored. To the best of our knowledge, only a few works have been dedicated to multiple FL services \cite{MultiTaskIOTJ2021,MyGC21,MultiFL-TWC,MultiFL-TMC}. Chen et al. \cite{MultiTaskIOTJ2021} formulated the client selection problem for multiple FL services as a many-to-many matching problem, aiming to reduce the overall latency of FL services. Similarly, Cheng et al. \cite{MyGC21} proposed a hypergraph-matching-based algorithm for multiple FL services in edge computing networks to maximize the overall FL service utility. Xu et al. \cite{MultiFL-TWC} studied a bandwidth allocation problem for multiple simultaneous FL services, where the clients of each FL service share the common bandwidth resources. Nguyen et al. \cite{MultiFL-TMC} investigated the resource-sharing problem for multiple FL services, and each client was assumed to participate in the training of multiple FL services simultaneously. However, these studies have not considered the client selection problem for multiple FLSDs with limited monetary budgets over dynamic candidate client sets. \textit{As far as we know, this paper is among the first to propose and design the data quality-aware dynamic client selection problem for multiple FLSDs under constrained monetary budgets.} Specifically, each client has multiple dynamic datasets and can participate in the training of multiple FL services simultaneously, while the FLSDs have to compete for clients with monetary incentives to reach their target training accuracies over multiple time rounds. The problem is formalized as a non-cooperative Markov game, where a multi-agent hybrid DRL (MAHDRL)-based algorithm is proposed to resolve the problem, via considering action conflicts avoidance between optimal client selection and payment actions. Besides, a data quality indicator (DQI) function is designed to prevent data-quality privacy disclosure. Extensive simulations on three datasets demonstrate the effectiveness of our proposed algorithm in comparison with representative algorithms.

\section{System model}
We consider a trading market of multiple FL services under the management of an FL platform (e.g., an edge server), which consists of a set of $M$ FLSDs, denoted by $\mathcal{M} = \{1,\dots,m,\dots, M\}$, and a set of $C$ clients, denoted by $\mathcal{C}=\{1,\dots,c,\dots,C\}$. Besides, the system time is slotted into $N$ time slots, denoted by $\mathcal{N} = \{1,\dots,n,\dots, N\}$. Each time slot $n$ can be regarded as a training round for each FLSD. Specifically, each FLSD $m \in \mathcal{M}$ represents a unique FL service task, and needs to recruit sufficient clients for iterative FL training at each time slot by considering their data size and distribution. Each client $c \in \mathcal{C}$ has multiple local datasets corresponding to $M$ FL services, which will be updated at the beginning of each time slot, and thus can guarantee the data freshness. For analytical simplicity, we assume that each time slot is long enough for the clients to finish one local training round. This assumption is reasonable as data collection and preprocessing usually take a long time.     Let $\mathcal{D}_{c,m}^{n}$ denote the local dataset of client $c$ that can be used to train FL service $m$ at slot $n$, where $D_{c,m}^{n} = |\mathcal{D}_{c,m}^{n}|$ represents the data size. Besides, the earth movers distance (EMD) value $\upsilon_{c,m}^{n}$ \cite{TMCAuction} is used to quantify the data distribution of $\mathcal{D}_{c,m}^{n}$. The smaller the value of EMD, the more the dataset's distribution tends to be IID. According to \cite{TMCAuction}, the EMD value of $\mathcal{D}_{c,m}^{n}$ can be calculated as,
\begin{equation}\label{e1}
\upsilon_{c,m}^{n}=\sum_{y_{c,m} \in \mathcal{Y}}\left\|\mathbb{P}_{c}(y_{c,m}^{n})-\mathbb{P}_{g}(y_{c,m}^{n})\right\|,
\end{equation}
where $\mathbb{P}_{c}$ denotes the actual data distribution of $\mathcal{D}_{c,m}^{n}$. $\mathbb{P}_{g}$ is a reference data distribution and can be shared as public knowledge over all the clients \cite{TMCAuction}. For example, regarding a multi-class classification problem, $y_{c,m}^n$ represents a class label.

Each client is supposed to be equipped with a $K$-core CPU. Thus, each client can participate in the training of multiple FL services simultaneously  (no more than $K$) \cite{MultiFL-TMC}. For analytical simplicity, suppose that clients can communicate with the FL platform directly, with stable wireless links. Besides, the overall number of clients stays unchanged; namely, no clients will leave/enter the market, while no new FLSDs will arrive within $T$ time slots. However, since FLSDs may have different convergence speeds, they will leave the trading market after service completion, e.g., reach a target accuracy.

Each FLSD $m$ aims to obtain a final trained model with a specific requirement, e.g., the target test accuracy. According to analysis associated with \cite{FedIncen-1}, \cite{TMCAuction} and \cite{MultiFL-TMC}: i) local training accuracy can be approximated via the overall data quality (i.e., data size, data distribution) of selected clients, and ii) global model can be optimized via optimizing the intermediate model in each round. Thus, our proposed client selection problem of FLSD $m$ can be regarded as a Markov decision process over multiple time slots. For simplicity, let 0-1 binary variable $x_{c,m}^{n} \in \vec{X}_{m}^{n}$ vector denote the client selection action of FLSD $m$ at time slot $n$, where $x_{c,m}^{n}=1$ indicates that client $c$ is selected by FLSD $m$; otherwise, $x_{c,m}^{n}=0$.

To motivate clients to participate in FL training, appropriate monetary reward, namely, a payment $P_{c,m}^{n} \in \vec{P}_{m}^{n}$ should be paid to the clients by FLSD $m$. More practically, the maximum payment that FLSD $m$ can pay at each slot should be limited, say, budget $B_{m}$\footnote{For simplicity, we assume the budget of each FLSD is the same at each time slot.}, which further limits the maximum number of clients that each FLSD can select \cite{FL-6G,DRLFL-CL22,TPDS-22}. Besides, each client claims a cost bid $O_{c,m}^{n}$ to the FL platform, which mainly accounts for the costs of data collection/storage, local model training, and communication \cite{TMCAuction}. To support clients' willingness, $P_{c,m}^{n}$ should be no less than $O_{c,m}^{n}$. Moreover, when client $c$ is selected by more than $K$ FLSDs, it can serve at most $K$ FLSDs who offer the highest payments above the corresponding cost bid. Given the limited CPU core, client number, and budget, we have the following key constraints,
\begin{equation}\label{e2}
0 \leq \sum_{m \in \mathcal{M}}x_{c,m}^{n} \leq K, \forall c \in \mathcal{C},
\end{equation}
\begin{equation}\label{e4}
0 \leq \sum_{c \in \mathcal{C}}x_{c,m}^{n}P_{c,m}^{n} \leq B_{m}, \forall m \in \mathcal{M},
\end{equation}
\begin{equation}\label{e5}
0 \leq O_{c,m}^{n} \leq P_{c,m}^{n}, \forall c \in \mathcal{C}, \forall m \in \mathcal{M}.
\end{equation}

\begin{figure}[t]
    \centering
    \subfigure[On MNIST]{\includegraphics[width=1.47in,angle=0]{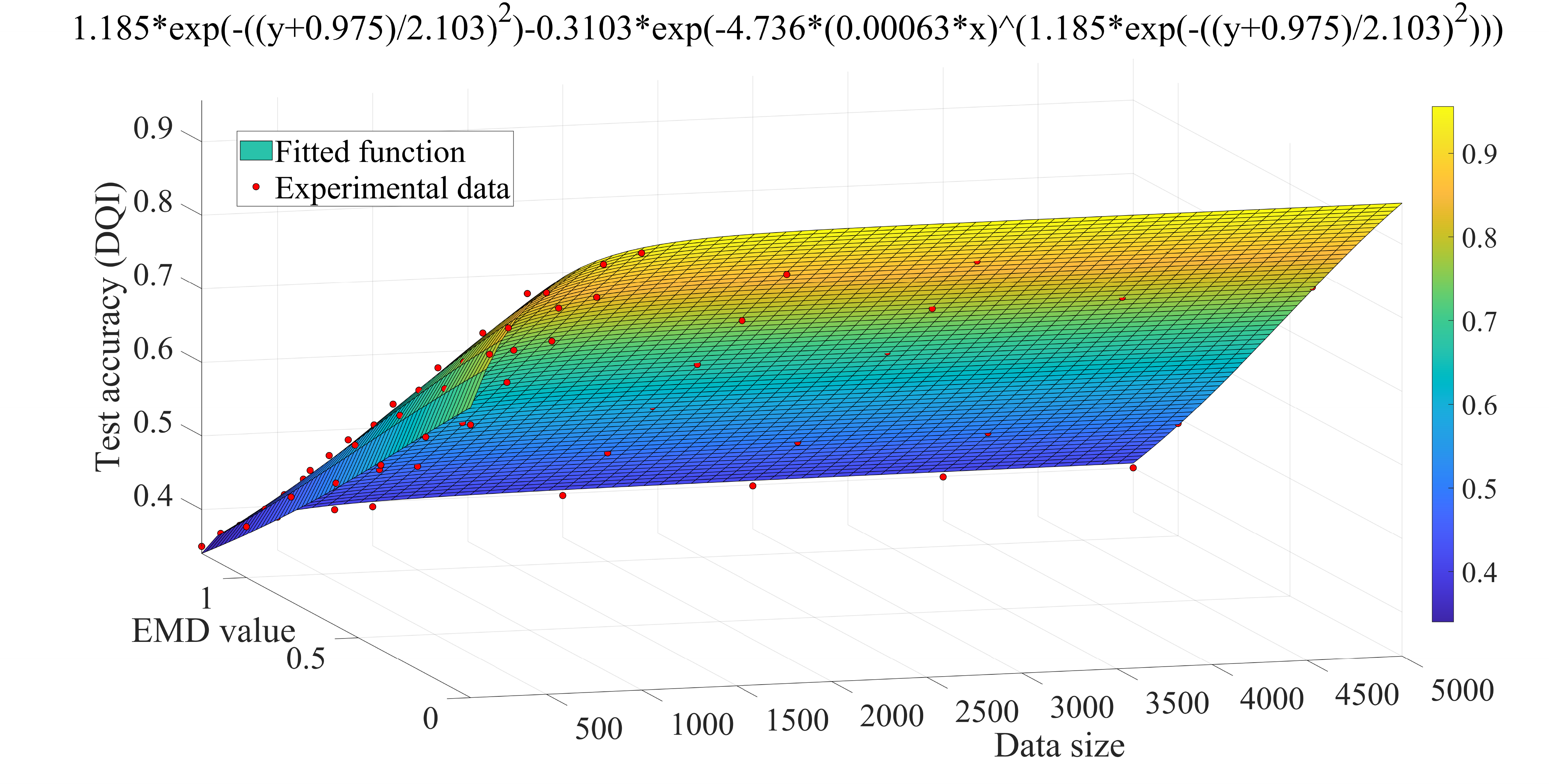}}
    \subfigure[On Fashion-MNIST]{\includegraphics[width=1.55in,angle=0]{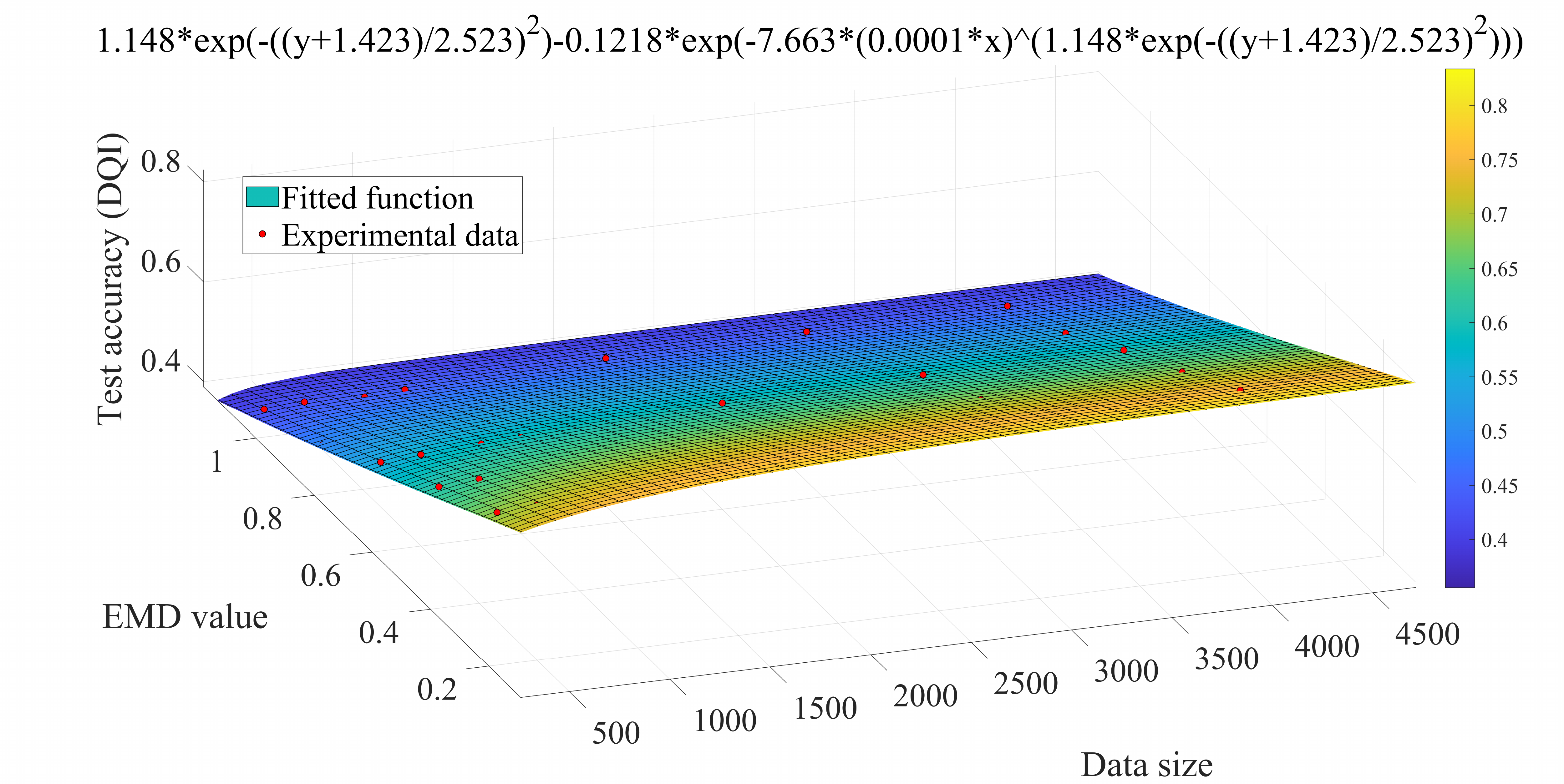}}
    \caption{The estimations of equation (5) on MNIST and Fashion-MNIST.}
    \label{fig1}
\end{figure}

\section{MADRL-based client selection}
In each time slot, each FLSD $m$ needs to decide both the client selection vector $\vec{X}_{c,m}^n$, and the corresponding payment vector $\vec{P}_{m}^n$ under constraints (\ref{e2})-(\ref{e5}), to hire feasible clients for FL training till the target test accuracy $\varpi_{m}$ has been reached. Considering dynamic datasets of clients and the giant selection space, conventional optimization methods such as problem decoupling and iterative optimization will face difficulties in handling dynamic optimizations, and also lead to high optimization costs. Thus,  DRL represents a promising solution. In the following, we formalize the problem as a non-cooperative Markov game, and a MAHDRL-based algorithm is proposed to resolve the problem.

\subsection{Markov game formation}
We define the Markov game by describing agents, state, action, and reward function for FLSDs as follows:
\begin{itemize}
  \item \textbf{Agents}: Assuming that $M$ FLSDs are independent and self-interested learning agents. Notably, the number of agents can be different from different time slots, since FLSDs exit the market after service completion.
  \item \textbf{State}: Client can report its local information to the FL platform at each slot $n$, including its data size vector $\vec{D}_{c}^{n}$, the corresponding EMD value vector $\vec{\upsilon}_{c}^{n}$, and the cost bid vector $\vec{O}_{c}^{n}$. However, data size and data distribution are privacy-sensitive information \cite{TPDS-22}, and clients may be reluctant to provide the information directly. To further prevent privacy disclosure, each client uploads a DQI value $\Psi_{c,m}^{n}$ instead. The basic idea of DQI value is to capture the relationship among the test accuracy, data size, and EMD value, and can be calculated by the following function (\ref{e6}), via adopting the function form proposed in \cite{TMCAuction},
    \begin{equation}\label{e6}
    \begin{aligned}
    \Psi_{c,m}^{n}=\alpha \left(\upsilon_{c,m}^{n}\right)-\eta_{1} \mathrm{e}^{-\eta_{2}\left(\eta_{3} D_{c,m}^{n}\right)^{\alpha \left(\upsilon_{c,m}^{n}\right)}},
    \end{aligned}
    \end{equation}
    where $\alpha \left(\upsilon_{c,m}^{n}\right)=\eta_{4} \exp \left(-\left(\frac{\upsilon_{c,m}^{n}+\eta_{5}}{\eta_{6}}\right)^{2}\right)<1$ is the function of EMD value, $\eta_{1},\eta_{2},\dots,\eta_{6}$ are curve fitting parameters. A larger DQI value means that the model can achieve higher accuracy as trained by the corresponding dataset. For example, we conduct extensive experiments to verify the effectiveness of (\ref{e6}) in Fig. (\ref{fig1}), which indicates that (\ref{e6}) can achieve a satisfying estimation of test accuracy, i.e., DQI, via data size and EMD value. In addition to protecting privacy, using DQI rather than data size and EMD value can reduce the communication cost and the dimensionality of state.

Then, the FL platform can generate a global information $G^{n}=\{ {\Psi}_{m}^{n},\mathbf{O}^{n}\}$ on behalf of the FLSDs, which is available to all the FLSDs. Thus, the state of FLSD $m$ at slot $n$ can be denoted as $s_{m}^{n}=\{G^{n},B_{m}\}$. Since $B_{m}$ represents a private information and unavailable to other FLSDs, each FLSD can only observe partial state of the market.
  \item \textbf{Action}: When each agent $m$ obtains the state $s_{m}^{n}$, it takes the action $a_{m}^{n}=\{\vec{X}_{m}^{n},\vec{P}_{m}^{n}\}$ to select feasible clients, where $\vec{X}_{m}^{n}$ represents the discrete client selection action, and $\vec{P}_{m}^{n}$ indicates a continuous payment action.
  \item \textbf{Reward}: Inspired by \cite{INFOCOM-20} and \cite{TPDS-22}, we define the instantaneous reward of FLSD $m$ at time slot $n$ as $r_{m}^{n}=\Omega_{m}^{\varpi_{m}^{n}}$, where $\varpi_{m}^{n}$ is the test accuracy after $t$ time slots, $\Omega_{m}\gg1$ is a constant that guarantees the reward grows exponentially with the test accuracy. Note that FLSD $m$ exits the market but still gets a constant reward (e.g., $\Omega_{m}^{\varpi_{m}}$) when $\varpi_{m}^{n}\geq\varpi_{m}$. By learning the client selection and payment actions, each FLSD $m$ aims to maximize the accumulated reward $R_{m}=\sum_{n=1}^{N}\gamma^{n-1}r_{m}^{n}$, where $0<\gamma \leq 1$ denotes a discount factor.
\end{itemize}

\begin{algorithm}[t]\label{Alg1}
\small
\caption{MAHDRL-Based Client Selection and Payment Algorithm}
\begin{algorithmic}[1]
		\State Initialize the Actor network $\mu_{m}^{\theta_{m}}$ and Critic network $Q^{\omega_{m}}$ for each agent $m \in \mathcal{M}$; Initialize the target networks $\mu_{m}^{\theta_{m}^{\prime}}$ and $Q^{\omega_{m}^{\prime}}$, with $\theta_{m}^{\prime}=\theta_{m}$, $\omega_{m}^{\prime}=\omega_{m}$.
        \State Initialized learning parameters: Learning episode $EP$, maximum steps per episode $N$, replay buffer $\mathcal{B}$, mini-batch size $B$, returned episode reward $\mathbf{R}$
        \FOR {Each agent $m \in \mathcal{M}$}
        \State Obtain an initialized state $s_{m}^{I}$
        \FOR{Each episode ep = 1:EP}
        \State Generate a random process $\xi$ for continuous action exploration; Set the total episode reward $R_{m}^{ep}=0$.
         \FOR{Each training step n = 1:N}
        \State Each agent $m$ obtains initial local state $s_{m}^{n}$, and further obtains the continuous action $a_{m,c}^{t}=\mu_{m}^{\theta_{m}}(s_{m}^{n})+\xi$.
        \State Obtain the discrete action $a_{m,d}^{n}$ via a preset policy (e.g., $\epsilon$-greedy).
        \State Obtain the combined action $(a_{m,d}^{n},a_{m,c}^{n})$; Execute the action $(a_{m,d}^{n},a_{m,c}^{n})$ and obtain the reward $r_{m}^{n}$, and the next state $s_{m}^{n+1}$.
        \State Store the tuple $(s_{m}^{n},a_{m,d}^{n},a_{m,c}^{n},s_{m}^{n+1})$ into the replay buffer $\mathcal{B}$.
        \State Set $R_{m}^{ep}+=r_{m}^{n}$.
        \State Sample some tuples from the replay-butter to make up a mini-batch.
        \State Update the Actor network according to (\ref{e9}), update the Critic network according to (\ref{e10}) and (\ref{e11}).
        \State Update the target networks according to (\ref{e12}).
        \ENDFOR
        \ENDFOR
        \State Set $\mathbf{R} \Longleftarrow R_{m}^{ep}$.
        \ENDFOR
\end{algorithmic}
\end{algorithm}

\subsection{MAHDRL-based solution}
According to the above discussions, two key problems exist during the design of the MADRL algorithm to address the proposed Markov game: dynamics of agents, and hybrid action space. The former problem can be addressed by adopting a fully independent MADRL framework where each FLSD learns independently \cite{UAV-TWC}, which can further ensure scalability and flexibility. The latter problem can be addressed by applying a MAHDRL algorithm, via combining deep q-network (DQN) and deep deterministic policy gradient (DDPG) \cite{P-DQN}. Specifically, the basic idea of HDRL is to parameterize the discrete action and figure out which discrete-continuous hybrid action pair can generate the maximum action value. Such action splitting can effectively keep the joint optimization gain, while avoiding the conflict between discrete and continuous actions. For analytical simplicity, we take agent $m$ as the representative of multiple agents, and assume that the corresponding action $a_{m}=\{a_{m,d},a_{m,c}\}$ is split into the discrete action $a_{m,d}=\vec{X}_{m}$ and the continuous action $a_{m,c}=\vec{P}_{m}$. Based on the action splitting, the \textit{Actor-Critic} architecture can be used. Suppose that the discrete action set of each agent is $\mathcal{H}=\{1,2,...,H\}$, where the size of $\mathcal{H}$ depends on the total number of all discrete action combinations. To further reduce the action space, the serial action selection method can be adopted as supported by \cite{INFOCOM-20} and \cite{TPDS-22}, which samples one client from the client set at a time, until exceeding the budget. For each $a_{m,d}\in \mathcal{H}$, the corresponding continuous action $a_{m,c}$ can be regarded as its associated parameters. Based on which, the Actor with a deterministic policy network $\mu_{m}^{\theta_{m}}({s}_{m})$ can be used to map the state to the continuous action $a_{m,c}=\mu_{m}({s}_{m})$. Meanwhile, the Critic is applied to generate the corresponding discrete action and evaluate the Actor's output action by outputting Q-values, with a Q-value function network $Q_{m}^{\omega_{m}}\left(s_{m},\left({a}_{m,d},{a}_{m,c}\right)\right)$, by taking $s_{m}$ and $a_{m,d}$ as input, which is parameterized by $\omega_{m}$. Notably, to stabilize the training performance, target networks are applied for the Actor and Critic, which are parameterized by $\theta_{m}^{\prime}$ and $\omega_{m}^{\prime}$, respectively. Then, the optimal discrete-continuous action pair can be selected by the following (\ref{e7}),
\begin{equation}\label{e7}
 \left(a_{m,d}^{*}, a_{m,c}^{*}\right)=\max \limits_{\left({a}_{m,d}, {a}_{m,c}\right)} Q_{m}^{\omega_{m}}\left(s_{m},\left({a}_{m,d},{a}_{m,c}\right)\right).
 \end{equation}

 Thus, for any given discrete action $a_{m,d}$, i.e., when the $\omega_{m}$ is fixed, we aim to seek for $\theta_{m}$ such that,
\begin{equation}\label{e8}
 Q_{m}^{\omega_{m}}\left(s_{m},{a}_{m,d},\mu_{m}^{\theta_{m}}\left({s}_{m}\right)\right)= \max\limits_{a_{m,d}\in [H]}Q_{m}^{\omega_{m}}(s_{m},a_{m,d},a_{m,c}).
 \end{equation}

According to the above discussions, similar to DDPG training \cite{P-DQN}, we can update the actor network (discrete action) by sampling $B$ data samples from the replay buffer and calculating the following gradient
\begin{equation}\label{e9}
\small
\begin{aligned}
&\nabla_{\theta_{m}} \mathcal{J}(\mu_{m}^{\theta}) \approx \\
&\left.\frac{1}{B} \sum_{b} \nabla_{\theta_{m}} \mu_{m}^{\theta_{m}}\left(s_{m}^{b}\right) \nabla_{a_{m,d}^{b}} Q_{m}^{\omega_{m}}\left(s_{m}^{b},{a}_{m,d}^{b},\mu_{m}^{\theta_{m}}({s}_{m}^{b})\right)\right.,
\end{aligned}
\end{equation}
where superscript $b$ indicates the time slot index.

Similar to DQN, the Critic network is updated by minimizing the following loss function,
\begin{equation}\label{e10}
\mathcal{L}\left(\omega_{m}\right)=\frac{1}{B} \sum_{b}\left(y_{m}^{b}-Q_{m}^{\omega_{m}}(s_{m}^{b},a_{m,d}^{b},a_{m,c}^{b})\right)^{2},
\end{equation}
\begin{equation}\label{e11}
y^{b}_{m}=r_{m}^{b}+\gamma Q_{m}^{\omega_{m}^{\prime}}(s_{m}^{b+1},a_{m,d}^{b+1},a_{m,c}^{b+1}).
\end{equation}

Finally, the target Actor and Critic networks are updated as follows,
\begin{equation}\label{e12}
\begin{array}{l}
\theta_{m}^{\prime} \leftarrow \varsigma \theta_{m}+(1-\varsigma) \theta_{m}^{\prime}, \\
\omega_{m}^{\prime} \leftarrow \varsigma \omega_{m}+(1-\varsigma) \omega_{m}^{\prime}.
\end{array}
\end{equation}
where $0<\varsigma\ll1$ is a constant. The overall training algorithm of the proposed MAHDRL is summarized in Algorithm 1. The convergence performance of Algorithm 1 can be analyzed in the same way as the Theorem 1 in \cite{UAV-TWC}, which is omitted here due to space constraint. As for the time complexity of Algorithm 1, it can be denoted as ${\cal O}\left( {{T_{max}}\left( {{N_{in}}{n_g} + 2\sum\limits_{g = 0}^{G - 1} {{n_g}} {n_{g + 1}}} \right) + 2{S^2}A + {{\log }_2}B} \right)$ for the training process according to \cite{MyIoT}, where $T_{max}$ is the maximum training step, $N_{in}$ is the input size of the Actor network and Critic network; ${n_g}$ is the size of each layer when assuming the Actor network and Critic network both have $G$ hidden layers. Besides, $S$ and $A$ are the size of state and action, respectively. $B$ is the size of replay buffer. Similarly, the execution delay of Algorithm 1 is ${\cal O}\left( {{N_{in}}{n_g} + 2\sum\limits_{g = 0}^{G - 1} {{n_g}} {n_{g + 1}}} \right)$.

\begin{figure}[htbp]
    \centering
    \subfigure[]{\includegraphics[width=1.3in,angle=0]{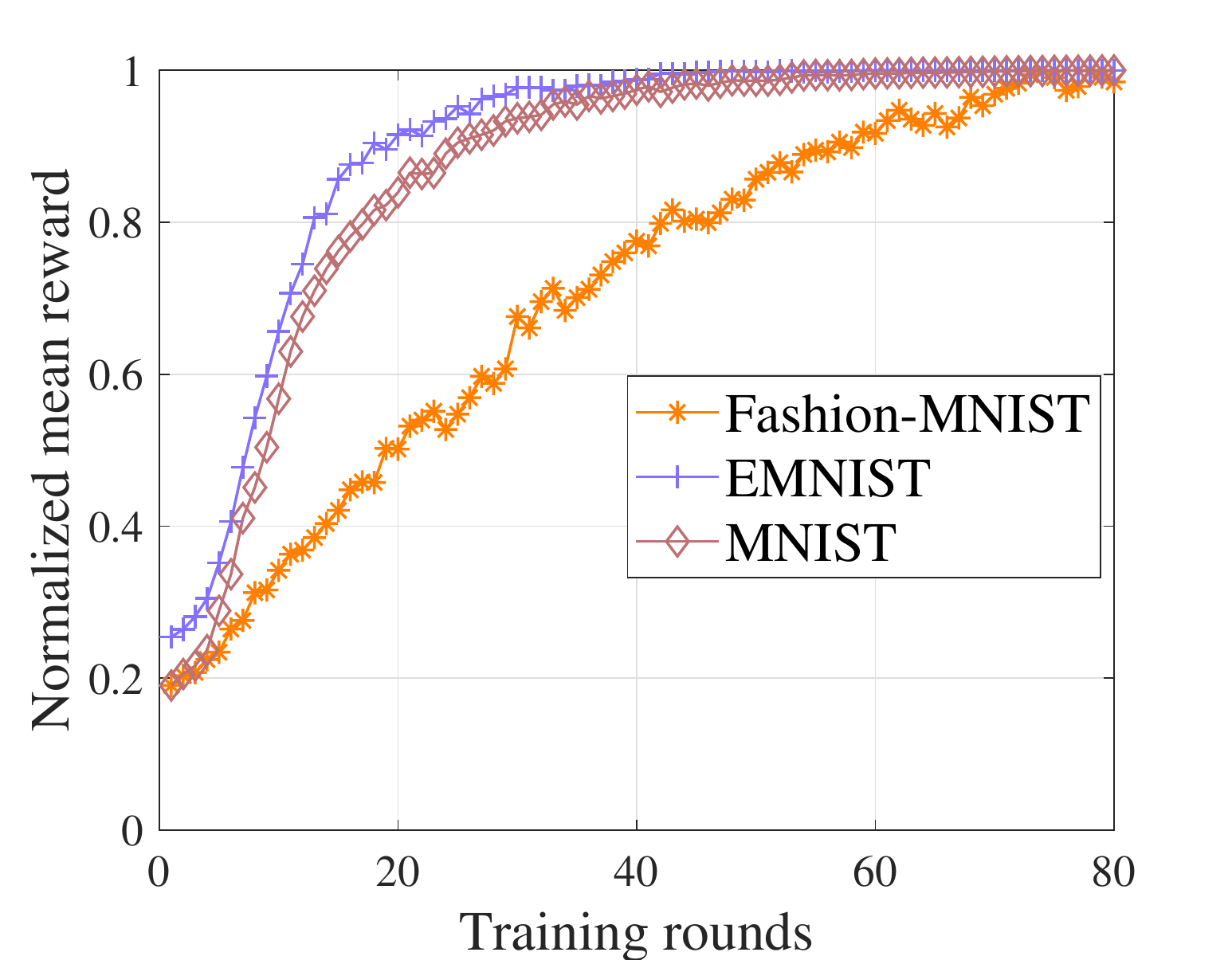}}
    \subfigure[] {\includegraphics[width=1.3in,angle=0]{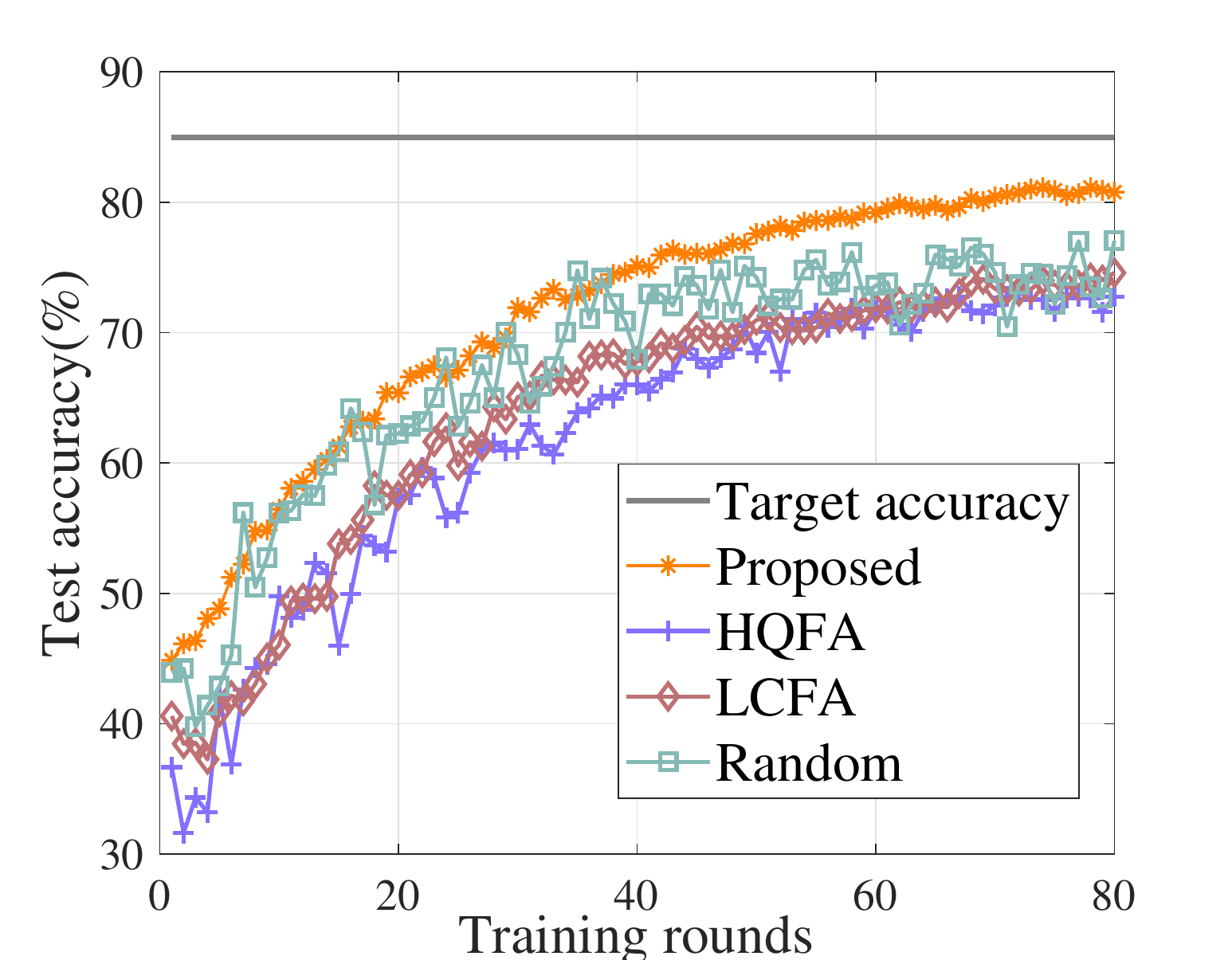}}
    \subfigure[] {\includegraphics[width=1.3in,angle=0]{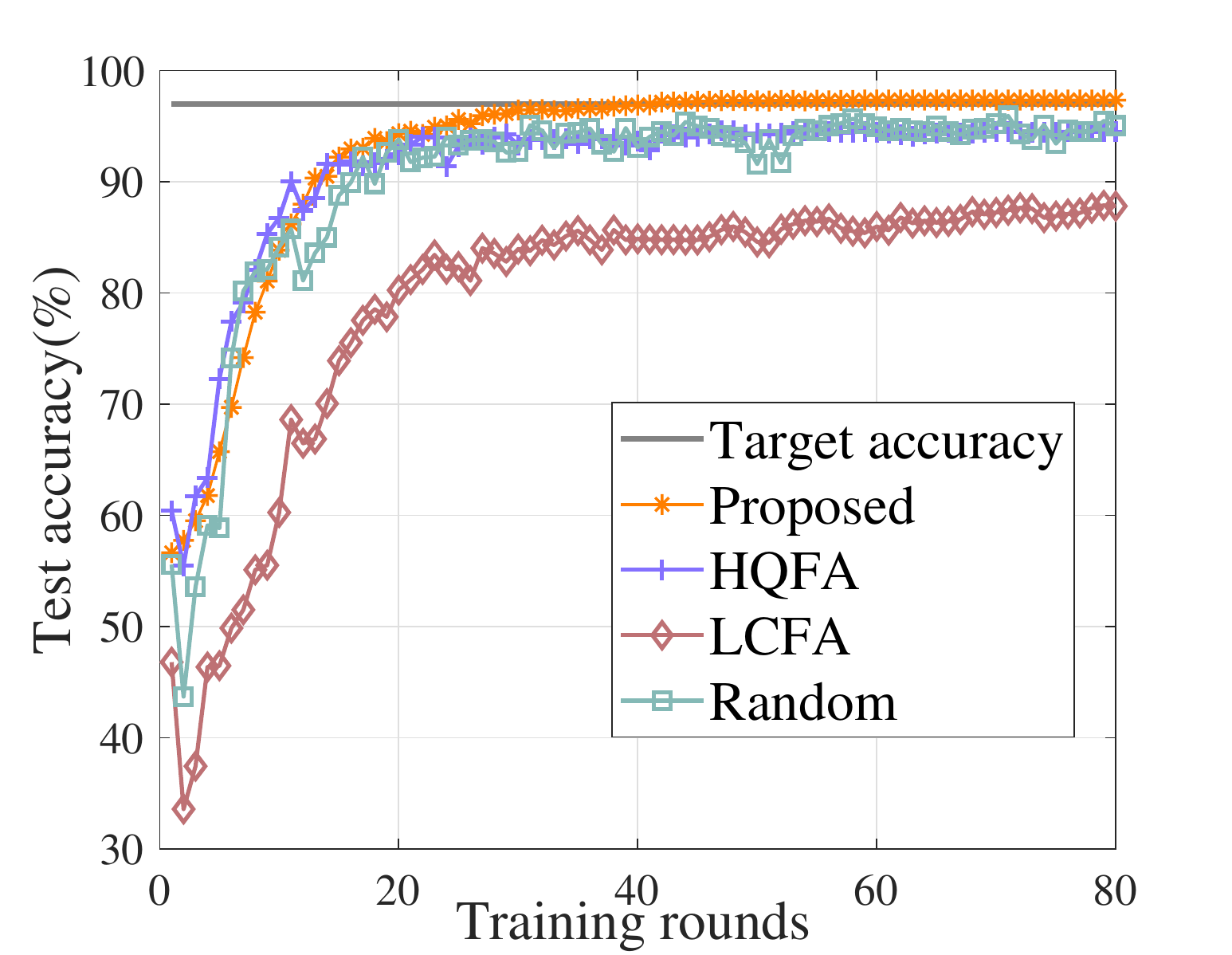}}
    \subfigure[] {\includegraphics[width=1.3in,angle=0]{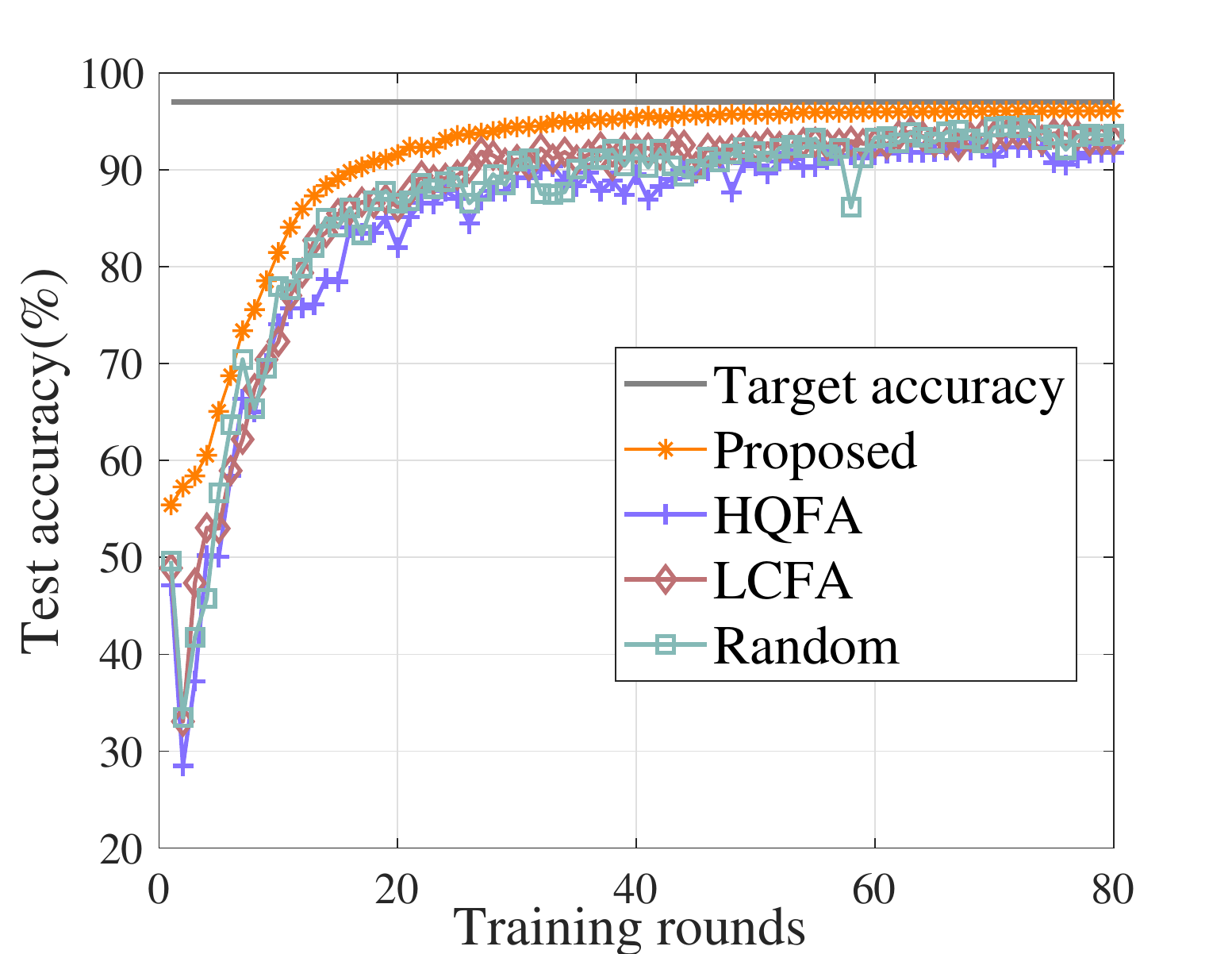}}
    \caption{Performance evaluations over the training rounds: (a) convergence performance; (b) test accuracy on Fashion-MNIST; (c) test accuracy on EMNIST; (d) test accuracy on MNIST.}
    \label{fig2}
\end{figure}

\section{Numerical Simulation}
\subsection{Simulation settings}
This section evaluates the proposed algorithm by considering three FL tasks (three FLSDs) on three datasets: i) MNIST, which is a dataset containing gray images of ten classes of handwritten digits ``0''-``9''; ii) Fashion-MNIST, which is a dataset containing gray images of ten classes of clothing; iii) EMNIST, which is a set of handwritten character digits derived from the NIST Special Database 19 and converted to a dataset structure that directly matches the MNIST dataset. A network model composed of two convolution layers and three full-connected layers is applied for all three datasets. The DQI curve fitting parameters of MNIST and Fashion-MNIST are shown in Fig. \ref{fig1}, while the parameters $\eta_{1}...\eta_{6}$ for EMNIST are set by -0.1922, 0.2613, 0.00063, 0.7084, 0.3189, 1.233, respectively. We assume there are 20 clients, each with two CPU cores (i.e., $K=2$) by default. Data size $D_{c,m}^{n}$ and EMD value $\upsilon_{c,m}^{n}$ of each client is selected from the set $\{100, 200,300, 400\}$ and $\{0.4, 0.6, 0.8, 1.0\}$, respectively. Then, the cost bid can be calculated by the data size and EMD value, as $O_{c,m}^{n}= 0.025*D_{c,m}^{n}-\upsilon_{c,m}^{n}$. Besides, the budget for each FLSD is set to 20. Considering the complexity of different datasets, the target accuracy on MNIST, Fashion-MNIST, and EMNIST are set to 97\%, 85\%, and 97\%, respectively, while $\Omega_{m}$ is set to 60, 100, and 30 for MNIST, Fashion-MNIST, and EMNIST, respectively. As for DRL settings, we conduct the simulation for 200 episodes and $80$ steps (training rounds) per episode. The replay buffer size is $4000$, and the batch size is 32. The discount factor $\gamma$ is set as 0.95, and $\varsigma=0.01$. The Actor network and Critic networks have two hidden layers comprised of 120 and 60 neurons, respectively.

Three representative algorithms from \cite{TPDS-22} are considered as benchmarks to evaluate the performance gain of our proposed algorithm, namely, Low-Cost First Algorithm (\textbf{LCFA}), High-Quality First Algorithm (\textbf{HQFA}), and \textbf{Random}. LCFA aims to match the FLSDs to the clients with the lowest cost under constraints (\ref{e2})-(\ref{e5}), thus can select as many clients as possible. HQFA matches the FLSDs to the clients with the highest DQI value under constraints (\ref{e2})-(\ref{e5}), which can thus select as many high-quality clients as possible. In Random, clients are randomly assigned to FLSDs under constraints (\ref{e2})-(\ref{e5}).

\subsection{Results analysis}
Fig. \ref{fig2} evaluates the training performance for different algorithms on three datasets over the training rounds. Specifically, Fig. \ref{fig2}(a) curves the normalized mean reward of the proposed MAHDRL algorithm for three FLSDs (datasets). We can see that the mean rewards of the three FLSDs increase rapidly with the progress of training, EMNIST and MNIST reach the convergence around the 40-th round, while Fashion-MNIST converges around the 70-th round, which proves the commendable convergence performance of our proposed algorithm. Fig. \ref{fig2}(b)-Fig. \ref{fig2}(d) illustrate the test accuracy performance for different algorithms. With the highest test accuracy, our proposed algorithm outperforms the other algorithms on three datasets, offering good robustness and obvious superiority. Interestingly, compared with LCFA and HQFA, Random can obtain similar or even higher test accuracy on the three datasets. According to the defined cost, the reason is that LCFA selects more low-quality clients so that the model performance cannot be optimized; while HQFA gives priority to clients with high data quality, a small number of selected clients also hinders the further improvement of the model.

\begin{figure}[htbp]
    \centering
    \subfigure[]{\includegraphics[width=1.3in,angle=0]{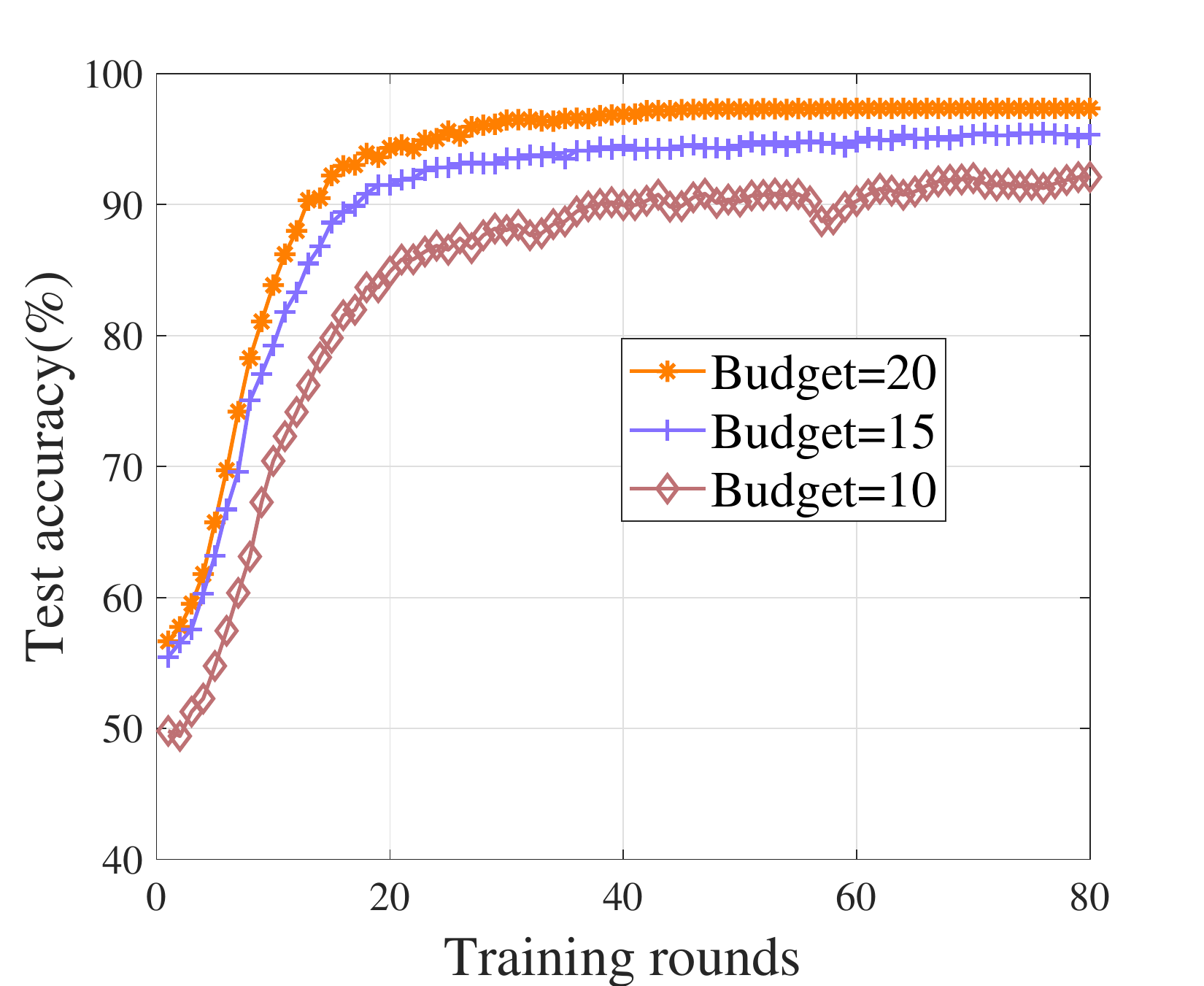}}
    \subfigure[] {\includegraphics[width=1.3in,angle=0]{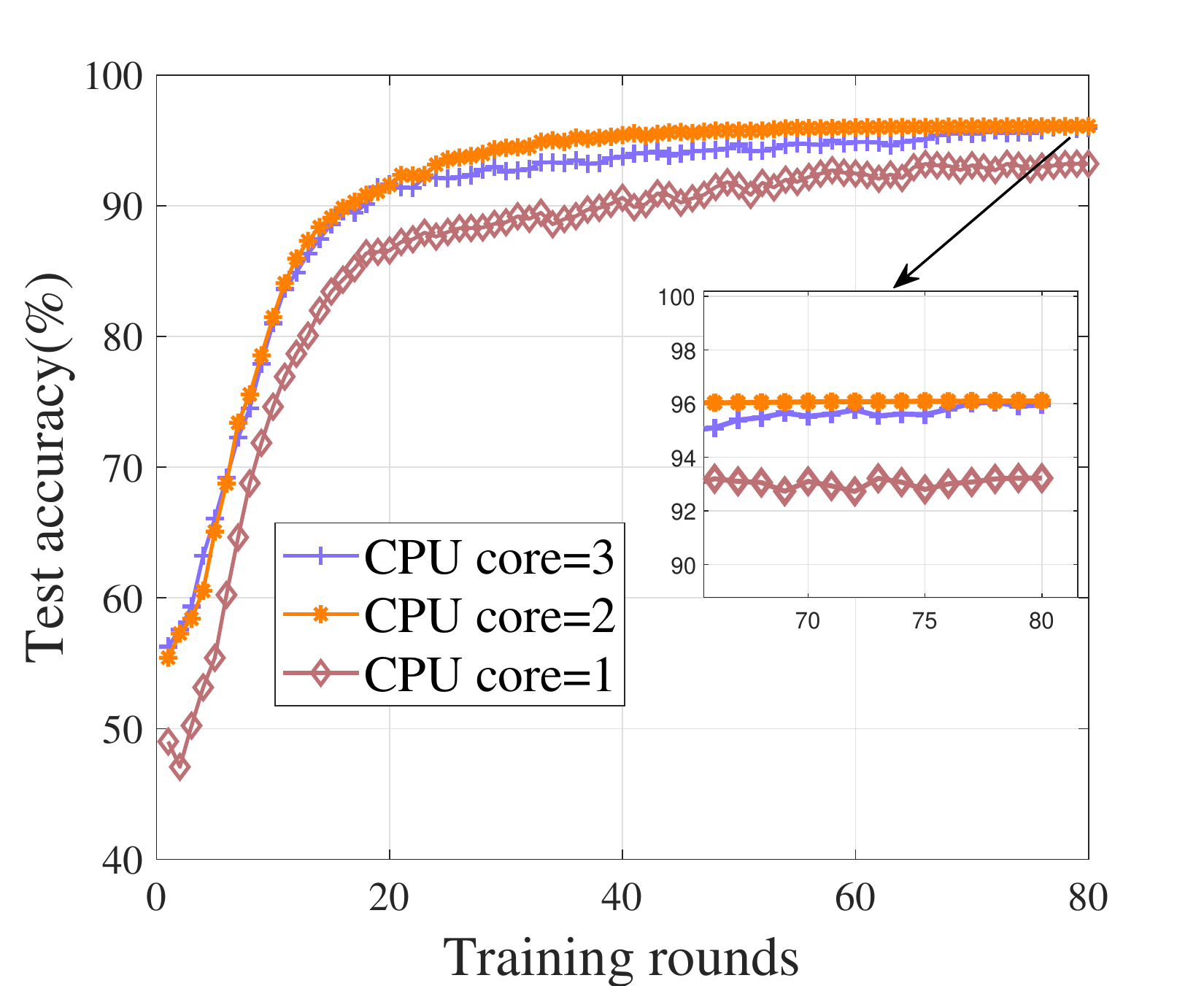}}
    \caption{Training performance comparisons for various budget and CPU core values: (a) test accuracy on EMNIST; (b) test accuracy on MNIST.}
    \label{fig3}
\end{figure}

Fig. \ref{fig3} compares the performance of the proposed algorithm upon considering different parameter settings. Fig. 3(a) shows the training performance on EMNIST when the corresponding budget equals 20, 15, and 10, respectively. A larger budget allows the FLSDs to select more high-quality clients, while achieving a better model performance, which is also consistent with the previous analysis. Fig. 3(b) depicts the training performance under a different number of CPU cores (i.e., $K$) on MNIST. Compared to $K=1$, increasing the number of CPU cores can achieve higher training performance since more  CPU cores can reach similar efforts by increasing the number of candidate clients, which also provides a more extensive search space. Although the same target test accuracy can be reached when the number of CPU cores equals to 2 and 3, a slower convergence speed can be incurred when that equals to 3 due to a larger searching space.

\section{Conclusion}
This paper investigates a quality-aware client selection problem for multiple FL services with constrained budgets in a wireless network. Specifically, each client can participate in the simultaneous training of multiple FL services, and each FLSD can select multiple clients while deciding the corresponding payments under budget constraints. The problem is formulated as a non-cooperative Markov game upon considering clients' dynamic data qualities and cost bids. We propose a MAHDRL-based algorithm to resolve the problem by learning the optimal client selection and payment actions for each FLSD. Simulation results on three practical datasets demonstrate that our proposed algorithm outperforms the representative algorithms significantly on training performance.

\section*{Acknowledgments}
The work presented in this paper was partially supported National Natural Science Foundation of China (Grant number 62271424, 61871339, 61971365), Key Science and technology Project of Fujian Province (Grant number 2020H6001) and Xiamen Major science and Technology Project (Grant number 3502Z20221026).

\section*{Conflict of interest}
The authors declare that there is no conflict of interest in this paper.



\bibliographystyle{elsarticle-num}

\end{document}